\documentclass{article}

\usepackage{natbib}

\usepackage{hyperref}       
\usepackage{url}            
\usepackage{booktabs}       
\usepackage{amsfonts}       
\usepackage{nicefrac}       
\usepackage{microtype}      
\usepackage{xcolor}         

\usepackage{amssymb}
\usepackage{amsmath,amscd}
\usepackage{amsthm}
\usepackage{cancel}     
\usepackage{centernot}  
\usepackage{graphicx}   
\usepackage{soul}       
\usepackage{color}      
\usepackage{enumitem}   
\usepackage{bbm}

\usepackage{algorithm}           
\usepackage{algorithmicx}        
\usepackage{algpseudocode}       

\usepackage{subcaption}
\usepackage{wrapfig}

\usepackage{fullpage}

\newcommand\numberthis{\addtocounter{equation}{1}\tag{\theequation}}

\newtheorem{thm}{Theorem}[section]
\newtheorem{lemma}[thm]{Lemma}

\newtheorem*{thm*}{Theorem}
\newtheorem*{lemma*}{Lemma}
\newtheorem*{cor*}{Corollary}
\newtheorem*{prop*}{Proposition}
\newtheorem*{conjecture*}{Conjecture}

\theoremstyle{definition}

\newtheorem*{defn*}{Definition}
\theoremstyle{definition}

\theoremstyle{definition}

\theoremstyle{remark}
\newtheorem*{ex*}{Example}
\theoremstyle{definition}

\theoremstyle{definition}
\newtheorem*{assm*}{Assumption}
\theoremstyle{remark}

\theoremstyle{remark}
\newtheorem*{remark*}{Remark}

\newcommand{\iid}{\overset{\text{iid}}{\sim}}
\newcommand{\R}{\mathbb{R}}
\newcommand{\E}{\mathbb{E}}

\newcommand{\sP}{\mathcal{P}}
\def\sC{\mathcal{C}}
\def\sX{\mathcal{X}}
\def\sY{\mathcal{Y}}

\def\tsX{\tilde{\sX}}

\def\hw{\hat{w}}
\def\hp{\hat{p}}

\def\kl{D_{\mathrm{KL}}}

\def\tO{\widetilde{O}}
\def\Xtest{X_{\text{test}}}
\def\ind{\mathbbm{1}}
\title{PMODE: Theoretically Grounded and Modular Mixture Modeling}

\author{
  Robert A. Vandermeulen\\
  \texttt{robert.anton.vandermeulen@gmail.com} 
}
\date{}

\begin{document}

\maketitle

\begin{abstract}
We introduce PMODE (Partitioned Mixture Of Density Estimators), a general and modular framework for mixture modeling with both parametric and nonparametric components. PMODE builds mixtures by partitioning the data and fitting separate estimators to each subset. It attains near-optimal rates for this estimator class and remains valid even when the mixture components come from different distribution families. As an application, we develop MV-PMODE, which scales a previously theoretical approach to high-dimensional density estimation to settings with thousands of dimensions. Despite its simplicity, it performs competitively against deep baselines on CIFAR-10 anomaly detection.
\end{abstract}

\section{Introduction}\label{sec:intro}
Mixture modeling is one of the most fundamental and flexible frameworks in statistical science, with core applications in clustering, density estimation, and latent-variable analysis. A canonical example is the Gaussian mixture model (GMM) \citep{yakowitz1968,bruni85,anderson14}, though numerous other forms—categorical, nonparametric, heterogeneous, etc.—have been studied extensively \citep{hall2003,elmore2005,vandermeulen2019operator,aragam2018npmix}. These models are particularly powerful when the data are generated by a latent process, such as in topic modeling \citep{blei01,arora12b,arora13,anandkumar14b,ritchie20}, where observations are conditionally independent given some hidden class or label. Mixture models are also widely used as flexible tools for density estimation, even if latent structure is not explicitly assumed.

Despite its ubiquity, mixture modeling poses significant statistical and computational challenges. For instance, until only a few years ago, the minimax-optimal rate for estimating the \emph{distribution} (as opposed to the parameters of the mixture components) of a Gaussian mixture was unknown \citep{ashtiani18-gaussian}. Moreover, algorithms that achieve this rate often involve solving computationally prohibitive optimization problems, particularly in high dimensions or under weak component separation. Bridging the gap between theoretical guarantees and practical methods remains a central challenge in the field.

A striking aspect of mixture models is that estimating a single density from the same class as the mixture components is often statistically and computationally straightforward. For example, estimating a single multivariate Gaussian at the minimax-optimal rate is well understood and computationally efficient. In contrast, learning a mixture of such distributions is substantially more difficult, especially in high dimensions or when the components are only weakly separated \citep{kwon20a}. This fact was leveraged in \cite{ashtiani18-mixtures}, where the authors assumed access to good estimators for the individual components—i.e., when given i.i.d. samples from a single component—and showed that these could be used to derive a general bound on the sample complexity of mixture models. While the resulting rate is not optimal in all cases, the bound applies under very broad conditions. To prove it, the authors introduced a concrete estimator, but one that requires solving a discrete optimization problem with no effective heuristics—not even for identifying a local minimum. As such, while the theoretical contribution is strong, the estimator itself offers little practical guidance for performing mixture modeling in applied settings.

We study a framework for mixture model estimation that generalizes the approach of \cite{ashtiani18-mixtures}, which we call the Partitioned Mixture Of Density Estimators (PMODE). We show that certain variants of PMODE—specifically those minimizing either $L^2$ distance or KL divergence—achieve estimation rates that are competitive with those in \cite{ashtiani18-mixtures}. While these estimators are computationally intractable in general, they naturally suggest heuristics for finding local optima in mixture modeling tasks. These methods retain the computational complexity of their base estimators, but also admit ``embarrassingly parallel'' optimization strategies.

To highlight PMODE's utility, we focus on high-dimensional density estimation, particularly in the context of multi-view modeling—a setting with strong theoretical guarantees for improved sample complexity in nonparametric mixture estimation \citep{vandermeulen2021,vandermeulen2023sample,chhor24}. We demonstrate that PMODE significantly improves the computational feasibility of multi-view approaches. Applied to the CIFAR-10 anomaly detection benchmark, our PMODE-based estimator outperforms a well-established deep image anomaly detection baseline, DSVDD \cite{ruff18}, on several classes, despite being a shallow method not specifically designed for image data.
 To our knowledge, this is the first successful application of multi-view modeling at this scale in both dimensionality and dataset size.

PMODE offers a general, modular, and principled framework for mixture modeling that is both theoretically grounded and practically effective.

\section{Background and Theoretical Method}

Mixture modeling is a widely used and deeply studied technique in statistics and machine learning. At its core, it aims to estimate an unknown distribution from data, typically assumed to take the form $p(x_1,\ldots,x_d) = \sum_{i=1}^k w_i p_i(x_1,\ldots,x_d)$,
where the \emph{component weights} $w_i \in \R$ are nonnegative and sum to one, and the \emph{components} $p_i$ are probability density functions (pdfs) drawn from some class of densities $\sP_i$. Most often, it is assumed that $\sP_i = \sP_j$ for all $i, j$, as in Gaussian mixture models, but there are important cases where the component classes differ—that is, $\sP_i \neq \sP_j$ for some $i \neq j$. These are known as \emph{heterogeneous mixture models}.

In this work, we focus exclusively on the problem of \emph{density estimation}—that is, estimating weights $\hw_i$ and densities $\hp_i$ from data such that the estimated mixture $\sum_{i=1}^k \hw_i \hp_i$ closely approximates the true mixture $\sum_{i=1}^k w_i p_i$. The problem of how individual components $w_i p_i$ correspond to components $\hw_j \hp_j$ in the estimator is an important and subtle one, but falls outside the scope of this paper \cite{ho16}.

Mixture models admit a natural generative interpretation, with each observation being generated in two steps. First, an unobserved component label $Y$ is sampled according to the weights $w = (w_1, \ldots, w_k)$, so that $P(Y = i) = w_i$; then, conditional on $Y = i$, an observation $X$ is drawn from the corresponding component distribution: $X \sim p_i$.

Given data $\sX_n = \{X_1, \ldots, X_n\} \iid p$, suppose that we had access to the (hidden) component labels $Y_1, \ldots, Y_n$ associated with each sample. In that case, the mixture modeling problem becomes trivial: for each class $j$, we could simply collect the data points with label $j$, denoted by $\sX_{n,j} = \{X_i : Y_i = j\}$, and apply a density estimator to this subset to produce an estimate $\hp_j$ of $p_j$.

A \emph{density estimator}, in our context, is any function that takes a dataset (e.g., $\sX_n$ or a subset such as $\sX_{n,j}$) and returns a probability density function. The most familiar example is perhaps the multivariate Gaussian estimator: given a set of samples, it computes the sample mean and covariance, and returns the corresponding Gaussian density. More sophisticated estimators include kernel density estimators (KDEs), which produce smooth, nonparametric densities. Crucially, we assume that any such estimator is fully specified—it accepts data as input and outputs a density. If hyperparameters (e.g., bandwidth in KDEs) are involved, we assume they are selected in a fixed, automatic way, such as via cross-validation or analytical rules such as Silverman's method for KDEs \citep{silverman86}, which depends only on the sample size $n$, the data variance, and the dimensionality $d$.

Returning to our main point: if the component labels were known, one could estimate each component density by applying a density estimator to the corresponding subset $\sX_{n,j}$, and estimate the component weights via $\hw_j = |\sX_{n,j}| / n$. The resulting mixture estimate would be $\sum_{i=1}^k \hw_i \hp_i$.

Of course, in practice the labels are not observed. One, perhaps simplistic, approach is to consider all possible assignments of labels to the data—that is, all possible partitions of the dataset into $k$ subsets—and evaluate each candidate partition using a hold-out set to estimate how well the resulting mixture approximates the true distribution under some loss function $\ell_n$ (e.g., empirical KL divergence, i.e., negative log-likelihood). We refer to this general approach as PMODE (Partitioned Mixture Of Density Estimators), and provide a summary in Algorithm~\ref{alg:pmode}.

\begin{algorithm}[hbt]
    \caption{Partitioned Mixture of Density Estimators (PMODE) - theoretical version}
    \label{alg:pmode}
    \begin{algorithmic}[1]
        \Require Training data $\mathcal{X}_n = \{X_i\}_{i=1}^n$, density estimators $V_1,\dots,V_k$, distance metric $\ell_n$, split ratio $s\in(0,1)$
        \Ensure Best assignment $S^\star$ of estimation data to components

        \State Let $m \gets \lfloor s n \rfloor$
        \State Randomly split $\mathcal{X}_n$ into:
        \begin{itemize}
            \item Estimation set $\tilde{\mathcal{X}} = \{X_i\}_{i=1}^m$ (size $m$)
            \item Validation set $\bar{\mathcal{X}} = \{X_i\}_{i=m+1}^n$ (size $n-m$)
        \end{itemize}

        \State $S^\star \gets \emptyset$, \ $L^\star \gets +\infty$

        \ForAll{$S \in [k]^m$}  \Comment{each $S=(i_1,\dots,i_m)$ assigns samples to components}
        \For{$j=1$ to $k$}
        \State $\tilde{\mathcal{X}}_j \gets \{X_i \in \tilde{\mathcal{X}} : S_i = j\}$
        \EndFor
        \State Define mixture estimator for $S$:
        \begin{equation*}
            \hat{p}_S(x) = \sum_{j=1}^k \frac{|\tilde{\mathcal{X}}_j|}{m} \; V_j(\tilde{\mathcal{X}}_j)(x)
        \end{equation*}
        \State Compute validation loss:
        \begin{equation*}
            L_S \gets \ell_n\bigl(\hat{p}_S, \, \bar{\mathcal{X}}\bigr)
        \end{equation*}
        \If{$L_S < L^\star$}
        \State $L^\star \gets L_S$, \ $S^\star \gets S$
        \EndIf
        \EndFor

        \State \Return $S^\star$  \Comment{optimal partition of estimation data for final estimator}
    \end{algorithmic}
\end{algorithm}
For clarity, we note that $\ell_n$ does not use all $n$ samples; the subscript simply indicates that it is an empirical estimate. Selecting the split ratio $s$—the proportion of data assigned to partitions for estimating components (as opposed to validation)—is a key design choice. Although it can be cross-validated, Theorems \ref{thm:l2-pmode} and \ref{thm:kl-pmode} also give theory-driven guidance for choosing $s$ as a function of $n$. Those bounds favor a \emph{small} $s$—conservative in that it leaves more data for validation and mitigates high-variance behavior—a preference we have also observed empirically. Developing principled, data-driven rules for $s$ remains an interesting direction for future work.

While PMODE was developed independently in this work, we note that related ideas have appeared in the literature, most notably in \cite{ashtiani18-mixtures}, which investigates a similar mixture estimation strategy but focuses exclusively on theoretical analysis. Our framework differs both in motivation and in scope: PMODE is designed to support arbitrary component estimators and heterogeneous component classes, and emphasizes modularity and generality in both theory and implementation.

At first glance, PMODE might appear sample-inefficient due to the need to evaluate a loss over a combinatorially large number of possible assignments $sn$ of samples to components. However, a simple union bound combined with concentration results like Hoeffding's inequality shows that, in fact, PMODE can still be sample-efficient. The key insight is that even though the number of candidate partitions is large, each individual candidate can be evaluated with high-probability guarantees, and the union bound scales well enough to ensure overall control of the error.

\subsection{Related Work}

Mixture modeling has been widely applied across statistical and machine learning domains. The most well-known method is \emph{expectation maximization} (EM) \cite{dempster77,benaglia09}, which assigns soft labels—probability distributions over components—to each data point. The literature occasionally refers to ``hard EM'' \citep{wen23}, where samples are assigned to a single component, but this approach is far less popular; soft EM remains the standard. 

Outside of EM, spectral methods have been proposed for both Gaussian mixtures \citep{anandkumar14b} and nonparametric mixtures \citep{hall2003, hall2005mixture, song14}. Other approaches treat mixture modeling as a form of clustering, either to initialize EM (as in \textsc{scikit-learn} \citep{scikit-learn}) or as part of a nonparametric modeling pipeline \citep{dan18, vandermeulen2019operator, vandermeulen2024, aragam20, vankadara21, aragam22, aragam23}. There is also interest in deep learning approaches to mixture modeling \citep{hyvarinen19, virolli19, khemakhem20}.

The work of \cite{ashtiani18-mixtures} analyzes an algorithm closely related to PMODE. They show that if each component of a mixture can be estimated at rate $n^{-1/q}$ in $L^1$, then the full mixture can be estimated at rate $\tO(n^{-1/(q+2)})$. We extend this result in Theorem~\ref{thm:main} to allow heterogeneous component families and estimators.

Let $\Delta^k$ denote the $k$-dimensional probability simplex (i.e., the set of nonnegative vectors summing to one), and let $[k] = \{1, 2, \ldots, k\}$. The notation $\tO$ refers to asymptotic bounds that suppress polylogarithmic factors. Theorem~\ref{thm:main} below presents our generalization; proofs for this work are deferred to Appendix~\ref{appx:proofs}.

\begin{thm}[Extension of \cite{ashtiani18-mixtures}]\label{thm:main}
Let $\sP_1, \ldots, \sP_r$ be families of distributions on $\mathbb{R}^d$, each equipped with an estimator $V_i$ that achieves a convergence rate of $\tO(n^{-1/q_i})$ in $L^1$ distance. Let $k \in \mathbb{N}$, and for each $j \in [k]$, let $I_j \subseteq [r]$ denote the set of allowable component families for the $j$-th mixture component.

Then, for any target density of the form
\[
    p = \sum_{j=1}^k w_j p_j, \quad \text{where } w \in \Delta^k \text{ and } p_j \in \bigcup_{\gamma \in I_j} \sP_\gamma,
\]
there exists an estimator that produces weights $\hat{w} \in \Delta^k$ and densities $\hat{p}_1, \ldots, \hat{p}_k$ such that
\[
    \left\| \sum_{i=1}^k \hat{w}_i \hat{p}_i - p \right\|_1 \in \tO\left(n^{-1/(2 + \max_i q_i)}\right),
\]
where each $\hat{p}_i$ is the output of an estimator $V_\gamma$ for some $\gamma \in I_i$, and both $\hat{w}$ and the $\hat{p}_i$ depend only on samples $X_1, \ldots, X_n \iid p$.
\end{thm}
Mixture models also raise a range of theoretical questions beyond the scope of this paper, including issues of identifiability \citep{teicher63, teicher1967, yakowitz1968, bruni85} and parameter estimation rates \citep{ho16, kivva22}. We refer to additional related work as it arises throughout the paper.

\subsection{Relationship to PMODE}\label{sec:relationship}

PMODE builds on the analysis of \cite{ashtiani18-mixtures}, but extends it in a key direction: we generalize the framework to loss functions—such as $L^2$ and KL divergence—that admit stable and tractable optimization heuristics. This makes the method more practical while still achieving convergence rates comparable to those obtained in prior work. For the remainder of this paper, we focus on the \emph{non-adaptive, heterogeneous} setting, where all components are drawn from known classes with corresponding estimators.

As mentioned earlier, the theoretical algorithm in \cite{ashtiani18-mixtures} is similar in spirit to PMODE, but it avoids direct loss evaluation. Instead, it relies on a comparison-based procedure using Scheffé estimators \citep{scheffe47, yatracos85, devroye01}. A Scheffé estimator takes two candidate densities \(f\) and \(g\) along with i.i.d. data from an unknown distribution \(p\), and returns either \(f\) or \(g\), say \(\hat{p}\), such that with high probability,
\begin{equation} \label{eqn:scheffe-opt}
    \left\| p - \hat{p} \right\|_1 \le 3 \min\left(\left\|p - f\right\|_1, \left\|p - g\right\|_1\right) + \varepsilon.
\end{equation}
A key limitation in using this to select an estimator that it is only guaranteed to distinguish between candidates when one is at least a factor of three worse than the other, making it ill-suited for discrete optimization over partitions, where progress typically relies on small, incremental improvements. In the PMODE setting, the candidates \(f\) and \(g\) correspond to estimators obtained from slightly different partitions—e.g., by reassigning a single data point—which generally results in only a small change in performance. The three factor in \eqref{eqn:scheffe-opt} is unimprovable unless the estimator is allowed to return mixtures of the candidates, in which case a factor of two can be achieved—but this too is known to be unimprovable in general \citep{bousquet19}. This suggests that fully generic, optimal strategies like \cite{ashtiani18-mixtures} are not compatible with practical discrete optimization heuristics.

Other practical difficulties arise as well: implementing a Scheffé estimator involves computing integrals over data-dependent regions with moving boundaries, complicating both evaluation and optimization. We discuss these issues in more detail in Appendix~\ref{appx:scheffe}. Given these challenges, it is more practical to use stable, directly computable metrics to evaluate candidate partitions. To this end, we focus on PMODE estimators based on distance metrics such as $L^2$ and KL divergence, both of which are analytically tractable and computationally feasible in our framework.
\subsection{\texorpdfstring{$L^2$}{TEXT}-PMODE}

We now consider a version of PMODE that selects the partition minimizing the estimated squared $L^2$ distance to the true distribution $p$. The squared $L^2$ distance between a candidate density $f$ and the target $p$ is:
\begin{equation*}
    \left\|f - p\right\|_2^2
    = \int \left( f(x) - p(x) \right)^2 dx
    = \int f(x)^2 dx - 2 \int f(x)p(x) dx + \int p(x)^2 dx.
\end{equation*}
The final term, $\int p(x)^2 dx$, is independent of $f$ and can be ignored for optimization.

The first term, $\int f(x)^2 dx$, is often analytically tractable—for example, when $f$ is a mixture of Gaussians. The cross term, $\int f(x)p(x) dx$, can be estimated from held-out data, $\bar\sX$:
\begin{equation}
    \int f(x)p(x) dx = \mathbb{E}_{X \sim p}[f(X)] \approx \frac{1}{n - m} \sum_{i = m+1}^n f(X_i),
\end{equation}
where $X_{m+1}, \ldots, X_n$ comprise the validation set. So one has $\ell_n(f) = -2\frac{1}{n - m} \sum_{i = m+1}^n f(X_i) + \left\|f\right\|_2^2$.

While $\int f(x)^2 dx$ can also be estimated from data, we find performance improves significantly when this term is computed exactly. For $L^2$-PMODE, we obtain the following convergence rate.

\begin{thm}\label{thm:l2-pmode}
    Let $\sP_1,\ldots,\sP_k$ be families of distributions on $\R^d$, each admitting an estimator $V_i$ that achieves a convergence rate of $\tO(n^{-1/q_i})$ in $L^2$ distance. Assume further that there exists a constant $C > 0$ such that all densities in $\sP_i$ and all densities in the range of $V_i$ are bounded above by $C$. Then, applying $L^2$-PMODE with split ratio $s = n^{-4/(\max_i q_i + 4)}$ yields an estimator, $V$, such that, for any target density $p = \sum_{i=1}^k w_i p_i$ with $p_i \in \sP_i$, one has
    \[
        \left\| p - V \right\|_2 \in \tO\left(n^{-1/(4 + \max_i q_i)}\right).
    \]
\end{thm}

\subsection{KL-PMODE}

We now consider a version of PMODE that uses the Kullback--Leibler (KL) divergence as the objective for selecting the optimal partition. KL divergence is a natural choice in statistical modeling, especially for density estimation, as minimizing it is equivalent to maximizing the likelihood of the observed data under the candidate model.

Given a candidate density $f$ and the true density $p$, the KL divergence is defined as:
\[
    \kl(p \,\|\, f) = \int p(x) \log\left( p(x)/f(x) \right) dx.
\]
Since $\int p(x) \log p(x) \, dx$, is independent of $f$, minimizing KL divergence reduces to minimizing the expected negative log-likelihood:
\[
    \mathbb{E}_{X \sim p}[-\log f(X)] \approx -\frac{1}{n - m} \sum_{i = m+1}^n \log f(X_i),
\]
where the approximation uses the held-out validation set $\bar{\mathcal{X}} = \{X_{m+1}, \ldots, X_n\}$.

KL-PMODE therefore selects the partition that minimizes empirical negative log-likelihood. This is especially appealing when the density estimators are probabilistically well-calibrated—that is, when their values meaningfully reflect likelihood—or when likelihood-based evaluation is standard, as in Gaussian mixtures or deep generative models. An additional advantage is that KL divergence naturally supports log-space computation, which improves numerical stability when evaluating products of densities in high-dimensional settings. For KL-PMODE, we obtain the following convergence rate.

\begin{thm}\label{thm:kl-pmode}
    Let $\sP_1, \ldots, \sP_k$ be families of densities supported on $[0,1]^d$, each equipped with an estimator $V_i$ satisfying the convergence rate $\kl(p \,\|\, V_i) = \tO(n^{-1/q_i})$ for $p \in \sP_i$. Assume further that there exist constants $C_L, C_U > 0$ such that all densities in $\sP_i$ and in the range of $V_i$ are bounded below by $C_L$ and above by $C_U$. Then, applying PMODE with split ratio $s = n^{-2/(\max_i q_i + 2)}$ yields an estimator, $V$, such that, for any target density $p = \sum_{i=1}^k w_i p_i$ with $p_i \in \sP_i$, one has
    \[
        \kl\left(p \,\middle\|\, V\right) \in \tO\left(n^{-1/(2 + \max_i q_i)}\right).
    \]
\end{thm}

\subsection{PMODE Variants and Practical Optimization}

The PMODE framework supports a wide range of instantiations, depending on the choice of distance metric and component density estimators. Each combination yields a different variant. Among the metrics we consider, $L^2$ distance offers robustness to outliers, as it does not heavily penalize low-probability assignments. KL divergence, in contrast, avoids the need to compute integrals and enables optimization in log space—a useful property when working with products of densities, as we will exploit in Section~\ref{sec:app}.

The main computational challenge in PMODE lies in searching over partitions. Exhaustively evaluating all possible assignments $S \in [k]^m$ in Algorithm~\ref{alg:pmode} is intractable for even moderately sized datasets, due to the exponential number of possible partitions. In practice, one must turn to \emph{discrete optimization} to find approximately optimal partitions.

Standard discrete optimization strategies—such as hill climbing, simulated annealing, and beam search—are well-suited to this setting. Since the PMODE objective is modular and each candidate can be evaluated independently, these approaches are \emph{embarrassingly parallelizable}, allowing multiple candidate partitions to be explored simultaneously.

\section{Application: High-Dimensional Nonparametric Density Estimation} \label{sec:app}

We now consider, in depth, an instantiation of PMODE tailored to the setting of high-dimensional nonparametric density estimation. While PMODE is a general and flexible framework, we view this particular application as one of the central contributions of this work. Additional experiments considering applying PMODE to Gaussian mixture models can be found in Appendix \ref{appx:supp-experiments}.

\subsection{Background and Related Work}
High-dimensional statistics is a vibrant area of research, and high-dimensional \emph{nonparametric} density estimation remains a particularly challenging problem. Even under mild regularity conditions—such as assuming the target density is Lipschitz continuous—the minimax optimal convergence rate for estimating a $d$-dimensional density is $n^{-1/(2 + d)}$ \citep{stone80,stone82,devroye1985nonparametric,devroye01,tsybakov2009introduction,mcdonald17}. This is a classic manifestation of the curse of dimensionality: as the dimension $d$ increases, the rate deteriorates rapidly.

One promising line of work aims to sidestep this curse using \emph{multi-view models}, sometimes referred to as ``naive Bayes mixtures'' or simply ``nonparametric mixture models'' \citep{hall2003,hall2005mixture,benaglia09,song13,song14,bonhomme16,kargas19,vandermeulen2021,vandermeulen2023sample,kwon21,amiridi22I,amiridi22II,chhor24}. These models assume the data-generating density takes the form
\begin{equation}\label{eqn:multiview}
    p(x_1,\ldots,x_d) = \sum_{i=1}^k w_i \, p_{i,1}(x_1) p_{i,2}(x_2) \cdots p_{i,d}(x_d),
\end{equation}
where each $p_{i,j}$ is a one-dimensional density, and the weights $w_i$ are nonnegative and sum to one. This can be interpreted as a mixture model whose components are \emph{separable} across dimensions, i.e., each component density factorizes as $p_i(x) = \prod_{j=1}^d p_{i,j}(x_j)$.

Densities satisfying the multi-view assumption in \eqref{eqn:multiview} can be estimated at rate $\tO(n^{-1/3})$—independent of the dimension $d$, number of components $k$, or Lipschitz constants—and this rate is achievable even without knowledge of $k$ or the Lipschitz parameters \citep{vandermeulen2021}. The work of \cite{chhor24} studies the case $d = 2$ in detail, including a wider space of smoothness assumptions, and provides an efficient algorithm that achieves this optimal rate of convergence. To our knowledge, it remains the only known algorithm that is both computationally tractable and provably rate-optimal in this setting.

A major bottleneck in many existing multi-view methods is the need to factorize high-dimensional histograms or empirical tensors. This step becomes computationally prohibitive as the data dimension $d$ increases, severely limiting scalability.

\subsection{MV-PMODE}\label{ssec:mvpmode}

We now apply PMODE to the setting of high-dimensional density estimation under the multi-view assumption, resulting in a scalable and practical variant we term \emph{MV-PMODE}.

Each component density in MV-PMODE is modeled as a product of univariate kernel density estimators (KDEs), forming a naive Bayes model. Let $Z_1, \ldots, Z_n$ be $d$-dimensional samples, where $Z_i = (Z_{i,1}, \ldots, Z_{i,d})$. (We use ``$Z$'' to indicate that these may be samples from a subset in one of the PMODE partitions.) The component density is defined as:
\[
    q(x_1, \ldots, x_d) = \prod_{j=1}^d q_j(x_j), \quad \text{where } q_j(x_j) = \frac{1}{n} \sum_{i=1}^n k_\sigma(x_j, Z_{i,j}),
\]
where $k_\sigma$ is the Gaussian kernel with bandwidth $\sigma$:
\[
    k_\sigma(x, y) = \left(\sigma\sqrt{2\pi}\right)^{-1} \exp\left( -(x - y)^2/(2\sigma^2) \right).
\]
The bandwidth $\sigma$ controls the degree of smoothing: larger values produce smoother estimates, while smaller values concentrate density more locally. As the number of samples increases, $\sigma$ should typically decrease. Though bandwidth selection is a deep topic, we adopt \emph{Silverman's rule of thumb} \citep{silverman86}, which sets $\sigma_i \propto \hat{\sigma}_i \cdot n^{-1/5}$ for each dimension $i$, where $\hat{\sigma}_i$ is the sample standard deviation in the $i$-th coordinate of the current partition. This is simple, computationally efficient, and empirically effective in high dimensions.

Bringing these components together, the MV-PMODE component estimator is defined as a product of $d$ univariate KDEs:
\begin{equation} \label{eqn:naive_bayes_kde}
    V_j\left(\tsX_j\right)(x) = \prod_{i=1}^d \Big( \left|\tsX_j \right|^{-1} \sum_{X_r \in \tsX_j} k_{\sigma_i}(x_i, X_{r,i}) \Big).
\end{equation}
Each dimension is treated independently, allowing the bandwidth $\sigma_i$ to adapt to the marginal variability. This flexibility is useful, as different dimensions may exhibit vastly different behavior.

While MV-PMODE could use other distance metrics, we focus on KL divergence, which is especially suitable in high dimensions due to its log-space formulation. Since component densities involve a product over $d$ terms, computing densities directly in high dimensions becomes problematic. KL divergence avoids this by enabling optimization in log space, where products become sums, significantly improving numerical stability. Our implementation supports this formulation and scales to thousands of dimensions and samples.

When the target density satisfies a naive Bayes structure, product-form bounds allow us to reduce multivariate error to a sum of univariate errors:
\begin{equation*}
\left\|\otimes_{i=1}^d f_i - \otimes_{i=1}^d g_i\right\|_1 \le \sum_{i=1}^d \left\|f_i - g_i\right\|_1, \quad \kl(\otimes_{i=1}^d f_i \|\otimes_{i=1}^d g_i) = \sum_{i=1}^d \kl(f_i \| g_i),
\end{equation*}
as shown in \cite{reiss89} (Lemma 3.3.7) and \cite{tsybakov2009introduction} (Section 2.3). A similar inequality holds for $L^2$ distance, assuming an upper bound on the densities.

Standard univariate estimators, such as KDEs applied to Lipschitz densities, achieve $\tO(n^{-1/3})$ rates in these metrics \citep{devroye1985nonparametric}. While MV-PMODE does not fully satisfy the assumptions of Theorem~\ref{thm:kl-pmode}—for instance, boundedness away from zero—it comes close: such conditions could be enforced with mild modifications, e.g., bounding support to $[0,1]^d$, using histograms, or assuming a lower bound. Similar adjustments would allow $L^2$-PMODE to meet the conditions of Theorem~\ref{thm:l2-pmode}.

Although MV-PMODE falls slightly outside the formal setting of our theory, its structure suggests it may still inherit favorable convergence behavior. In particular, Theorem~\ref{thm:kl-pmode} implies that MV-PMODE could plausibly achieve dimension-independent rates on the order of $\tO(n^{-1/5})$, which—while slower than the $\tO(n^{-1/3})$ rate in \cite{vandermeulen2021, chhor24}—remain attractive in high-dimensional, practical settings.

\subsection{Comparison to Other Multi-view Models}

Our approach to multi-view modeling differs substantially from prior work. Most existing methods represent the joint density using discretized histograms, which are then factorized to uncover structure. In two dimensions, for example, a histogram with $b$ bins per axis corresponds to a $b \times b$ matrix of nonnegative entries summing to one. In higher dimensions, this generalizes to tensors, and modeling proceeds by attempting to factor these tensors into low-rank components.

While convenient for low-rank approximation, this discretized view introduces two key limitations. First, histogram methods suffer from bin edge effects and discontinuities, which degrade estimation quality. KDEs avoid these issues, providing smoother and more accurate estimates. Second, global tensor factorization often obscures the local structure of individual components, especially when the underlying distribution is truly separable.

MV-PMODE takes a different approach. Each component is modeled independently using smooth one-dimensional KDEs, then combined via a product structure. This design enables far better scalability computationally.

More broadly, PMODE’s modular structure offers significant flexibility. In MV-PMODE, each marginal can be estimated independently, allowing improvements to one-dimensional estimators to directly benefit the full model. This makes it easy to experiment with alternative estimators or distance metrics—such as replacing KDEs with histograms or even Gaussians—without reworking the entire framework.

This modularity also brings theoretical benefits: variants of PMODE inherit both the statistical guarantees of the base estimators and the convergence rates of the general framework. Algorithmically, PMODE avoids expensive joint training by enabling embarrassingly parallel optimization over candidate partitions, unlike tensor factorization methods that typically follow a fixed sequential path.

The method proposed by \cite{benaglia09} is perhaps closest in spirit to ours, using an EM-style algorithm to estimate multi-view densities with KDEs. However, it lacks the modularity, scalability, and generality of PMODE.

In summary, MV-PMODE combines theoretical soundness, scalability, and modularity in a unified framework—making it a practical and extensible approach to high-dimensional density estimation under multi-view assumptions.

\subsection{Anomaly Detection Experiment}\label{ssec:experiment}

Anomaly detection involves determining whether a test sample $\Xtest$ is drawn from the same distribution as a training set $X_1, \ldots, X_n \iid p$. A common approach is to estimate the density $\hp$ and classify a sample as anomalous if $\hp(\Xtest) \le \alpha$ for some threshold $\alpha$.

We evaluate MV-PMODE on the CIFAR-10 one-vs-rest anomaly detection benchmark. In this setting, one class is used as nominal (non-anomalous) training data, and the full test set is used for evaluation, with the remaining nine classes treated as anomalies. Performance is measured using AUROC. Full experimental details and results are in Appendix~\ref{appx:ad-experiment}; all experiments used $k = 20$ mixture components and were repeated five times per class.

Unlike many density estimation works that benchmark on synthetic or low-dimensional data and reimplement competing methods, we evaluate MV-PMODE on a high-dimensional vision benchmark using published results from the original authors. The CIFAR-10 one-vs-rest benchmark has been studied extensively for nearly a decade and remains a widely used testbed for image-based anomaly detection\footnote{See \url{https://paperswithcode.com/sota/anomaly-detection-on-one-class-cifar-10}}. Our comparisons use author-reported results from well-cited papers copied verbatim, ensuring consistency and avoiding artifacts from reimplementation.

The first baseline is Deep SVDD (DSVDD) \citep{ruff18}, a foundational deep method and widely used benchmark. The second is ADGAN \citep{deecke19}, a generative adversarial approach. We also include a naive Bayes baseline using univariate KDEs with Silverman's rule. Notably, all baselines are general-purpose rather than techniques highly engineered for this specific benchmark, such as \cite{tack20}. See Appendix~\ref{appx:image-ad-background} for background on the task and evaluation practices.

See Table \ref{tab:results} for results. While MV-PMODE does not achieve state-of-the-art results, unlike other nonparametric density methods, it competes directly with established deep methods on a challenging vision benchmark.
\begin{table}[ht]
    \centering \footnotesize
    \caption{Results for the CIFAR 10 one vs. rest anomaly detection test, AUC values. DSVDD \citep{ruff18} and ADGAN \citep{deecke19} are taken from their respective papers. Naive Bayes was performed via a KDE using Silverman's rule for each dimension. For MV-PMODE the average standard deviation was $0.69 \pm 0.41$, taken over the per class performance.}
    \label{tab:results}
    \begin{tabular}{l*{11}{c}}
        Method  & Air.  & Auto. & Bird  & Cat   & Deer  & Dog   & Frog  & Horse & Ship & Truck & Mean\\ \hline
        DSVDD    & 61.7  & \textbf{65.9}  & 50.8  & \textbf{59.1}   & 60.9  & \textbf{65.7}   & 67.7  & \textbf{67.3} & 75.9 & \textbf{73.1} &\textbf{64.8}\\
        ADGAN   & 66.1  & 43.5  & 63.6  & 48.8   & \textbf{79.4}  & 64.0   & 68.5  & 55.9 & \textbf{79.8} & 64.3 & 63.4\\
        Naive Bayes   & 65.8  & 45.8  & 62.0  & 50.9   & 74.3  & 52.6   & 71.3  & 55.5 & 70.3 & 57.9 & 60.6\\
        \underline{MV-PMODE}  & \textbf{73.8} & 48.9 & \textbf{68.8} & 51.3  &76.7  &50.5  &\textbf{75.3}  &54.6  & 75.0  & 54.0  &  62.9
    \end{tabular} 
\end{table}
\subsection{Discussion}\label{ssec:discussion}
\begin{figure}
    \centering
    \begin{subfigure}[b]{0.24\linewidth}
        \includegraphics[width=\linewidth]{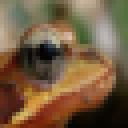}
        \caption{Frog (original)}
        \label{fig:frog}
    \end{subfigure}
    \begin{subfigure}[b]{0.24\linewidth}
        \includegraphics[width=\linewidth]{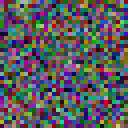}
        \caption{Frog (permuted)}
        \label{fig:frog-permuted}
    \end{subfigure}
    \begin{subfigure}[b]{0.24\linewidth}
        \includegraphics[width=\linewidth]{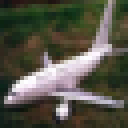}
        \caption{Airplane (original)}
        \label{fig:airplane}
    \end{subfigure}
    \begin{subfigure}[b]{0.24\linewidth}
        \includegraphics[width=\linewidth]{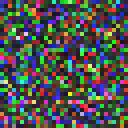}
        \caption{Airplane (permuted)}
        \label{fig:airplane-permuted}
    \end{subfigure}
    \caption{Examples of original and permuted images. The same permutation was used for both images.}
    \label{fig:frog-airplane}
\end{figure}Although MV-PMODE is a shallow, nonparametric density estimator not designed for anomaly detection, it outperforms influential deep methods on three CIFAR-10 classes by a meaningful margin. Notably, it ignores spatial structure: permuting pixel positions—including across color channels—has no effect on its performance, unlike deep models, which typically rely on convolutional architectures that exploit spatial locality. (To our knowledge, no transformer-based anomaly detectors operate directly on raw pixels without pretraining on external data.) The permutation clearly disrupts spatial coherence, making the image much harder to interpret by eye—an effect illustrated in Figure~\ref{fig:frog-airplane}. This structure-independence suggests a natural extension: combining MV-PMODE with nonparametric estimators that incorporate spatial structure \citep{vandermeulen24breaking,vandermeulen2025dimensionindependent}. Altogether, these observations underscore the potential of PMODE's modular design, which opens the door to a wide range of extensions and domain-specific adaptations.

\bibliography{refs.bib}
\bibliographystyle{plainnat}
\appendix
\section{Proofs}\label{appx:proofs}
This section presents the proofs of the theoretical results from the main text. The proof of Theorem~2.1 provides the core structure used throughout. We give it in full detail, while the proofs of Theorems~2.2 and~2.3 follow the same outline, and we include only the steps that differ.
\subsection{Proof of Theorem 2.1}
The following proof closely follows the argument in \cite{ashtiani18-mixtures}. The main contribution of Theorem~2.1 is to extend those results to the heterogeneous and adaptive settings. We begin with the following lemma.

\begin{lemma}[Theorem 3.4 page 7 of \cite{ashtiani18-gaussian}, Theorem 6.3 page 54 of~\cite{devroye01}] \label{lem:estimator-algorithm}
    There exists a deterministic algorithm that, given a collection of distributions $p_1,\ldots,p_M$, a parameter $\varepsilon >0$ and at least $\frac{\log \left(3M^2/\delta \right)}{2\varepsilon^2}$ iid samples from an unknown distribution $p$, outputs an index $j\in \left[M\right]$ such that
      \begin{align*} 
        \left\Vert p_j - p \right\Vert_1 \le 3 \min_{i \in \left[M\right]} \left\Vert p_i - p\right\Vert_1 + 4 \varepsilon
      \end{align*}
  with probability at least $1 - \frac{\delta}{3}$.
\end{lemma}

\begin{proof}[Proof of Theorem~2.1]
Let $q = \max_i q_i$ for brevity. Let $\sY_{1:n} = Y_1,\ldots, Y_n \iid w = [w_1,\ldots, w_k]$ (being interpreted as a discrete distribution) and let $X_i \sim p_{Y_i}$. From this it follows that $X_1,\ldots, X_n \iid p$. We split the sample of size $n$ into two parts:  
\[
    n_1 \;=\; \Bigl\lfloor n^\frac{q}{q+2} \Bigr\rfloor
  \quad\text{and}\quad 
  n_2 \;=\; n - n_1 \quad\text{ noting } n_2 = O(n).
\]
Denote the sequence corresponding to the first part of the data by $\sX_{1:n_1}$ (of length $n_1$) and the second part by $\sX_{n_1+1:n}$ (of length $n_2$). Note that we will use analagous colon notation for $\sY$ later. Let $k' = n_1$

\medskip
\noindent
\textbf{Step 1: Enumerating candidate mixtures.}
\smallskip

Consider \emph{all} ways to:
\begin{itemize}
    \item partition the $n_1$ samples in $\sX_{1:n_1}$ into at most $k$ nonempty blocks $B_1,\ldots,B_k$ (upper bounded by $k^{n_1}$ ways), and
    \item assign each block to one of the $s$ estimators $V_1,\ldots,V_s$ (upper bounded by $s^{k}$ ways).
\end{itemize}
Call each such labeled partition $c$; let the collection of them be $\sC$, with size $M = |\sC| \le k^{n_1}s^k.$  

For each $c \in \sC$:
\begin{itemize}
  \item If block $B \subseteq \{1,\dots,n_1\}$ is assigned the estimator $V_i$, then run $V_i$ on those $|B|$ samples (the corresponding subsample of $\mathbf{Y}$) to get $\widehat{p}_B$.  
  \item Let $\lambda_B := \tfrac{|B|}{n_1}$ be the empirical fraction.  
  \item Define the candidate mixture
  \[
    p_c \;=\; \sum_{B \subseteq c}\; \lambda_B \,\widehat{p}_B.  
  \]
\end{itemize}
Thus we produce a finite collection of mixture \emph{estimators} $\{p_c : c\in\sC\}$, with cardinality $k^{n_1}s^k$.

\medskip
\noindent
\textbf{Step 2: One candidate is close to the true mixture $p$.}
\smallskip

Among all our enumerations in $\sC$, there is a partition $c^*$ whose blocks exactly coincide with $\sY_{1:n_1}$ and which assigns block $B$ to the ``correct'' estimator $V_{a_j}$ (where $p_j\in\sP_{a_j}$).

Consider the block $B$ of $c^*$ that came from component $p_j$.  Since $V_{a_j}$ learns any $p_j \in \sP_{a_j}$ at rate $\widetilde{O}\bigl(m^{-1/q_{a_j}}\bigr)$, running $V_{a_j}$ on $|B|$ samples yields
\[
  \|\widehat{p}_B - p_j\|_{1} 
  \;\in\; 
  \widetilde{O}\!\Bigl(|B|^{-\tfrac{1}{q_{a_j}}}\Bigr).
\]
Also, $\lambda_B = |B|/n_1$ is close (within about $n_1^{-1/2}$, by Hoeffding’s or Chernoff bounds) to the true mixing weight $w_j$.  Since $|B|\approx w_j\,n_1$, the above implies
\[
   \|\widehat{p}_B - p_j\|_1 
   \;\in\; 
   \widetilde{O}\Bigl(\bigl(w_j\,n_1\bigr)^{-\tfrac{1}{q_{a_j}}}\Bigr)
   \;=\;
   \widetilde{O}\bigl(n_1^{-\tfrac{1}{q}}\bigr),
\]
because $q_{a_j}\le q$.  

Hence, summing over the $k$ blocks of $c^*$ (each assigned to the right $V_{a_j}$) and weighting each block by $\lambda_B$, it follows that 
\begin{align*}
  \|p_{c^*} - p\|_1
  & = \left\| \sum_{i=1}^k w_i p_i - \sum_{i=1}^k \lambda_{B_i} \hp_{B_i} \right\|_1\\
  & \le
  \sum_{i=1}^k \left|w_i -  \lambda_{B_i} \right| + \bigl\|\hp_{B_i} - p_i\bigr\|_1
  \;\in\;
  \widetilde{O}\bigl(n_1^{-\tfrac{1}{2}}\bigr)
  \;+\;
  \widetilde{O}\bigl(n_1^{-\tfrac{1}{q}}\bigr).
\end{align*}
Hence 
\[
  \min_{c\in \sC} \|p_c - p\|_1
  \;\le\;
  \|p_{c^*} - p\|_1
  \;\in\;
  \widetilde{O}\Bigl(n_1^{-\tfrac{1}{2}}\Bigr)
  + 
  \widetilde{O}\Bigl(n_1^{-\tfrac{1}{q}}\Bigr).
\]

\medskip
\noindent
\textbf{Step 3: Selecting the best candidate via Lemma~\ref{lem:estimator-algorithm}.}
\smallskip

We now apply Lemma~\ref{lem:estimator-algorithm} to choose \emph{which} $p_c$ best fits the second sample of size $n_2$.  Let $\varepsilon = n^{-1/(2+q)}\log(n)$ and let $\delta = 1/n$ in the algorithm with $n_2$ samples being used to choose the estimator. Recalling that $M= k^{n_1}s^k$ and $n_1 = O(n^{q/(q+2)})$, it is easy to verify that, for sufficiently large $n$, $n_2 \ge \tfrac{\log(3 M^2/\delta)}{2\,\varepsilon^2}$ and thus Lemma~\ref{lem:estimator-algorithm} ensures an output $\widehat{c}$ such that
\[
  \|\,p_{\widehat{c}} - p\|_1
  \;\le\;
  3 \,\min_{c \in \sC} \|p_c - p\|_1
  \;+\;
  4\,\varepsilon.
\]
with probability going to 1.
To see this observe that for sufficiently large $n$ there exists $C>0$ such that
\begin{align*}
    \frac{\log(3M^2/\delta)}{2 \varepsilon^2}
    & \le C O(n^{q/(2+q)})/ \left(\log^2(n) n^{-2/(2+q)} \right)\\
    & \le Cn/\log^2(n),
\end{align*}
which is less than $n_2$ for sufficiently large $n$, thus satisfying the assumptions of the lemma. Hence
\[
  \|\,p_{\widehat{c}} - p\|_1 \in
  \widetilde{O}\bigl(n^{-\tfrac{1}{2+q}}\bigr).
\]
\end{proof}
\subsection{PMODE Proofs}
\begin{proof}[Proof of Theorem 2.2]
    This proof proceeds similarly to that of the proof of Theorem 2.1. Letting $s =n^{-4/(q+4)} $ from the theorem statement, let $n_1 = sn = n^{q/(q+4)}$ and let $n_2 = n - n_1 = O(n)$. Let $\hp_n$ be the density selected by $L^2$-PMODE with $n$ samples.

\textbf{Step 1: Enumerating all candidate mixtures.}
\smallskip

Assigning each of the $n_1$ samples to one of the $k$ component estimators yields at most $k^{n_1}$ labeled partitions.

\textbf{Step 2: One candidate is close to the true mixture $p$.}
\smallskip

Proceeding as in the proof of Theorem 2.1 we have that
\begin{align*}
  \|p_{c^*} - p\|_2
  & = \left\| \sum_{i=1}^k w_i p_i - \sum_{i=1}^k \lambda_{B_i} \hp_{B_i} \right\|_2\\
  & \le \left\| \sum_{i=1}^k w_i p_i - \sum_{i=1}^k \lambda_{B_i} p_i\right\|_2 + \left\|\sum_{i=1}^k \lambda_{B_i} p_i - \sum_{i=1}^k \lambda_{B_i} \hp_{B_i} \right\|_2\\
  & \le \sum_{i=1}^k|w_i- \lambda_{B_i}| \left\|   p_i\right\|_2 + \sum_{i=1}^k\lambda_{B_i}\left\|  p_i  - \hp_{B_i} \right\|_2.\numberthis \label{eqn:l2-bias}
\end{align*}
Looking at the left term in \eqref{eqn:l2-bias}, using H\"older's Inequality we get that,
\begin{equation*}
    \left\|p_i \right\|_2 = \sqrt{\left\|p_i \right\|_2^2} = \sqrt{\left\|p_i\cdot p_i \right\|_1} \le \sqrt{\left\|p_i\right\|_1 \left\|p_i \right\|_\infty}  \le \sqrt{C},
\end{equation*}
and thus the left term has rate $O(n_1^{-1/2})$. The right term has rate $\tO(n_1^{-1/q})$ from the theorem hypotheses.

\textbf{Step 3: Selecting the best candidate via empirical risk minimization}
\smallskip

    Let $\ell(f) = -2 \left<f,p\right> + \left\|f\right\|_2^2$ and let $\ell_n(f)$ denote its empirical counterpart, as defined in the main text. Now we have that
    \begin{align*}
        \left\|p - \hp_n \right\|_2^2
        &=\left\|p - \hp_n \right\|_2^2 - \left\|p - p_{c^*}\right\|_2^2+ \left\|p - p_{c^*}\right\|_2^2\\
        & = \ell(\hp_n) - \ell(p_{c^*}) + \tO(n_1^{-2/q})\\
        & = \ell(\hp_n) -\ell_n(\hp_n) + \ell_n(\hp_n)- \ell_n(p_{c^*}) + \ell_n(p_{c^*})- \ell(p_{c^*})+ \tO(n_1^{-2/q})\\
        &  \le 2\sup_{c \in \sC} \left|\ell(p_c) -\ell_n(p_c)\right| + \tO(n_1^{-2/q})\\
        &  = 4\sup_{c \in \sC} \left|\frac{1}{n_2}\sum_{i=n_1 +1}^{n} p_c(X_i)-\E_{X\sim p}\left[p_c(X) \right]\right| + \tO(n^{-2/(q+4)}).
    \end{align*}
    From Hoeffding's inequality and the union bound we have that
    \begin{equation}
        P\left(\sup_{c \in \sC} \left|\frac{1}{n_2}\sum_{i=n_1 +1}^{n} p_c(X_i)-\E_{X\sim p}\left[p_c(X) \right]\right|\ge t \right) \le k^{n_1} \exp\left(\frac{-2n_2t^2}{C^2}\right). \label{eqn:l2-hoeffdings}
    \end{equation}
    Letting $t = \log(n) n^{-2/(q+4)}$, we get
    \begin{align}
        k^{n_1} \exp\left(\frac{-2n_2t^2}{C^2}\right) =  \exp\left(\log(k) n_1 - \frac{ -2n_2t^2}{C^2}\right),
    \end{align}
    and for the interior of the exponential, for sufficiently large $n$ there exists an $a>0$ such that
    \begin{align}
        \log(k) n_1 -\frac{2n_2t^2}{C^2}
        \le \log(k)n^{q/(q+4)}- a\frac{ n \log(n)n^{-4/(q+4)}}{C^2} \to -\infty,
    \end{align}
    and thus $\left\|p - \hp_n\right\|_2^2 = \tO(n^{-2/(q+4)})$, so $\left\|p - \hp_n\right\|_2 = \tO(n^{-1/(q+4)})$, which finishes the proof.

\end{proof}

\begin{proof}[Proof of Theorem 2.3]
    This proof proceeds similarly to that of the proof of Theorem 2.1 and 2.2. Letting $s = n^{-2/(q+2)}$ from the theorem statement, let $n_1 = sn = n^{q/(q+2)}$ and $n_2 = n - n_1 = O(n)$. Let $\hp_n$ be the density selected by KL-PMODE with $n$ samples.

\textbf{Step 1: Enumerating candidate mixtures.}
\smallskip

Partitioning $n_1$ into all possible partitions gives $k^{n_1}$ partitions.

\textbf{Step 2: One candidate is close to the true mixture $p$.}
\smallskip

We have that 
\begin{equation}
    \kl(p \| p^*_n) = \kl\left(\sum_{i=1}^k w_i p_i\| \sum_{i=1}^k \lambda_{B_i} \hp_{B_i}\right) \label{eqn:kl-bias-mixture}.
\end{equation}

The following lemma relates the KL-divergence of two mixtures:
\begin{lemma}[\cite{goldberger05,nielsen17}]
    Let $f = \sum_{i=1}^k \alpha_i f_i$ and $g = \sum_{i=1}^k \beta_i g_i$ be two mixtures of probability density functions, then
    \begin{equation}
        \kl(f \| g) \le \kl(\alpha \| \beta) + \sum_{i=1}^k \alpha_i \kl(f_i \| g_i).
    \end{equation}
\end{lemma}
Applying this to \eqref{eqn:kl-bias-mixture} we have
\begin{align*}
    \kl(p \| p^*_n) 
         &\le \kl\left( w \| \lambda \right) +  \sum_{i=1}^k \kl\left(p_i\| \hp_{B_i}\right)\\
         &= \tO(n_1^{-1/2}) + \tO(n_1^{-1/q})\\
         &= \tO(n^{-q/(2q+4)}) + \tO(n^{-1/(q+2)})\\
         &= \tO(n^{-1/(q+2)}).
\end{align*}
\textbf{Step 3: Selecting the best candidate via empirical risk minimization}
\smallskip

Let $\ell(f) =  -\E_{X\sim p} \log(f(X))$ and $\ell_n(f) = - \sum_{i=n_1+1}^{n} \log(f(X_i))$. We now have
    \begin{align*}
        \kl(p \| \hp_n) &= \kl(p \| \hp_n) - \kl(p \| p^*_n) +\kl(p \| p^*_n) \notag \\
                        &=\ell(\hp_n) - \ell(p^*_n) +\tO(n^{-1/(q+2)})\notag\\
                        &=\ell(\hp_n) - \ell_n(\hp_n) +\ell_n(\hp_n) -  \ell_n(p^*_n) + \ell_n(p^*_n) - \ell(p^*_n) + \tO(n^{-1/(q+2)})\notag\\
                        &\le \sup_{c \in \sC} 2\left|\ell(p_c) - \ell_n(p_c)\right| + \tO(n^{-1/(q+2)})\label{eqn:kl-bias-var}.\numberthis
    \end{align*}
Applying Hoeffding's inequality and the union bound, and using the assumed upper and lower bounds on the target densities and estimators, we find that there exists a constant $C > 0$ such that
    \begin{equation*}
        P\left(\sup_{c \in \sC} \left|\frac{1}{n_2}\sum_{i=n_1 +1}^{n} \log(p_c(X_i))-\E_{X\sim p}\left[\log(p_c(X)) \right]\right|\ge t \right) \le k^{n_1} \exp\left(\frac{-2n_2t^2}{C^2}\right). \label{eqn:kl-hoeffdings}
    \end{equation*}
    Letting $t = \log(n) n^{-1/(q+2)}$, we get
    \begin{align*}
        k^{n_1} \exp\left(\frac{-2n_2t^2}{C^2}\right) =  \exp\left(\log(k) n_1 -\frac{2n_2t^2}{C^2}\right).
    \end{align*}
    Evaluating the expression inside the exponential, we obtain, for sufficiently large $n$ and some $a>0$,
    \begin{align}
        \log(k) n_1 -\frac{2n_2t^2}{C^2}
        \le\log(k)n^{q/(q+2)}- a\frac{ n \log(n)n^{-2/(q+2)}}{C^2} \to -\infty,
    \end{align}
    and thus $\sup_{c \in \sC}\left|\ell(p_c) - \ell_n(p_c)\right| = \tO(n^{-1/(q+2)})$ and, from \eqref{eqn:kl-bias-var}, $\kl\left(p \| \hp_n\right) = \tO(n^{-1/(q+2)})$, which finishes the proof.

\end{proof}
\section{More on Scheff\'e Estimators} \label{appx:scheffe}

Consider the following problem: given two candidate density estimates $f$ and $g$, both independent of samples $X_1,\ldots,X_n \iid p$, determine which of the two is closer to the target density $p$ in total variation (i.e., $L^1$) distance. While this problem may appear straightforward, it is not immediately obvious how to design a statistically sound procedure for solving it. One elegant solution is the \emph{Scheff\'e estimator}, which we describe below.

Define the region $A = \left\{x \mid f(x) > g(x)\right\}$. The Scheff\'e estimator selects between $f$ and $g$ by comparing the discrepancy between their mass over $A$ and the empirical mass observed in the sample:
\begin{equation}\label{eqn:scheffe}
    \hp = 
    \begin{cases}
        f:& \left| \int_A f(x) dx - \frac{1}{n}\sum_{i=1}^n \ind(X_i \in A)\right| < \left| \int_A g(x) dx - \frac{1}{n}\sum_{i=1}^n \ind(X_i \in A)\right| \\
        g:& \text{otherwise}.
    \end{cases}
\end{equation}
The following result quantifies the performance of this estimator:
\begin{thm}[\cite{devroye01}, Theorem 6.1]
    For the estimator in \eqref{eqn:scheffe}, one has
    \begin{equation}\label{eqn:scheffe-theorem}
        \left\|p - \hat{p}\right\|_1 \le 3 \min\left(\left\|p - f\right\|_1,\left\|p - g\right\|_1 \right) + 4 \max_{B \in \mathcal{B}} \left|\int_B p(x) dx -\frac{1}{n}\sum_{i=1}^n \ind(X_i \in B) \right|,
    \end{equation}
    where $\mathcal{B} = \left\{\{x \mid f(x) > g(x)\}, \{x \mid f(x) < g(x)\} \right\}$.
\end{thm}
The second term in \eqref{eqn:scheffe-theorem} vanishes as $n \to \infty$ by the law of large numbers. The first term implies that the selected estimator is guaranteed to be within a factor of 3 of the best candidate in $L^1$ distance—often referred to as \emph{$3$-learnability}. To extend the procedure to more than two candidates, one can perform pairwise comparisons across all estimators, using the Scheff\'e estimators to identify the best via repeated tests. This approach, known as a \emph{Scheff\'e Tournament} (see \cite{devroye01}, Section 6.6), is statistically efficient: the union bound and Hoeffding’s inequality control error rates across all comparisons.

However, full Scheff'e Tournaments are typically computationally intensive, requiring $\binom{M}{2}$ pairwise comparisons for $M$ candidate densities. Some methods reduce this cost to linear in $M$ (e.g., \cite{mahalanabis07}), but in many applications—including \cite{ashtiani18-mixtures, ashtiani18-gaussian} and our own Theorem~2.1—the number of candidates grows exponentially or at least combinatorially with the sample size. As a result, the theoretical version of PMODE, like these prior approaches, remains computationally intractable at large scale.

As discussed in the main text, the factor of $3$ in \eqref{eqn:scheffe-theorem} makes it difficult to apply the Scheffé estimator in a principled way when selecting among a large number of candidate models. It is only guaranteed to distinguish between two candidates when one is at least a factor of three worse than the other in $L^1$ distance, making it poorly suited for optimization over partitions where typical improvements are incremental. Nonetheless, practitioners may still consider using \eqref{eqn:scheffe} as a heuristic, particularly when only a small number of candidates are being compared.

If the candidate densities $f$ and $g$ can be evaluated efficiently, computing the empirical term $\frac{1}{n} \sum_{i=1}^n \ind(X_i \in A)$, where $A = {x : f(x) > g(x)}$, is straightforward. However, evaluating the integral $\int_A f(x),dx$ poses a greater challenge. Since $A$ is defined implicitly and may be highly irregular, this integral is generally intractable in high dimensions. One workaround is to approximate it stochastically. Assuming the support of $f$ and $g$ lies within a known set $Q$, we can write:
\begin{align*}
\int_A f(x),dx
&= |Q| \int_Q \tfrac{1}{|Q|} f(x) \ind(f(x) > g(x)),dx \\
&= |Q| , \E_{Z \sim \mathrm{Unif}(Q)}\left[ f(Z)\ind(f(Z) > g(Z)) \right],
\end{align*}
where $|Q|$ is the Lebesgue measure of $Q$.

This expectation can be estimated via Monte Carlo, but in practice, we found the convergence rate $n^{-1/2}$ to be prohibitively slow. Even with millions of samples, the resulting estimates remained extremely noisy. Similar difficulties arose when attempting to estimate $\int f(x)^2 dx$ for $L^2$-PMODE. While Monte Carlo integration is theoretically viable, in practice the estimator tends to be extremely noisy—requiring an impractically large number of samples to achieve reliable accuracy—making it computationally inefficient and generally unsuitable for effective use.

\section{Background on CIFAR-10 One vs. Rest Benchmark} \label{appx:image-ad-background}

To our knowledge, the CIFAR-10 one-vs-rest benchmark (which we will shorten to ``CIFAR-10 AD'') was first introduced in \cite{ruff18}, and has since appeared in numerous anomaly detection papers.\footnote{See \url{https://paperswithcode.com/sota/anomaly-detection-on-one-class-cifar-10}} The benchmark has notable vulnerabilities: for instance, it can be nearly solved without training by using a CLIP model to compute the distance between an image embedding and a prompt such as “this is an image of an airplane,” achieving around 98.5 AUC \citep{liznerski2022exposing}. Prior to this, earlier methods proposed using auxiliary datasets (e.g., CIFAR-100) as source anomalies, a technique known as \emph{outlier exposure} \citep{hendrycks2018deep}. While these auxiliary datasets do not explicitly contain test-time anomaly classes, they almost certainly include samples visually similar to them. Indeed, \cite{liznerski2022exposing} showed that simply training a classifier to distinguish the nominal class from such an outlier dataset performs extremely well—even with just a few hundred outlier examples.

For methods that use \emph{no} external data—either implicitly or explicitly—the most effective approaches typically apply semantically meaningful transformations (e.g., rotations or shifts) to the data and train a network to predict the applied transformation \citep{golan18}. This is often done using a simple classification architecture, potentially with multiple output heads for composite transforms \citep{hendrycks19}. At test time, the network is applied to transformed versions of a test image, and the sample is declared anomalous if the model exhibits low confidence (e.g., low softmax score) in identifying the correct transformation. These scores can be aggregated across transformations. 

While effective, these methods depend heavily on prior knowledge about the data’s structure. For example, they assume that objects are typically centered and upright—assumptions that are true for CIFAR-10 but may not generalize elsewhere.

Among methods that rely solely on transformation prediction and use no external data—including pretrained networks—the strongest on CIFAR-10 AD is CSI \citep{tack20}, which achieves an impressive 94.3 AUC. CSI combines contrastive learning with transformation prediction, separating transformations into two groups: ``distribution-shifting'' transformations, treated as distinct classes in the contrastive loss; and ``shifting'' transformations, treated as non-contrasting and used as classification targets in an auxiliary network. The model is trained with a combination of these two objectives, and the final anomaly score blends the resulting outputs. While CSI is a well-engineered method, it is highly tailored to image data and specific to this benchmark, relying on dataset-specific assumptions and manual engineering.

For our experiment, we chose to focus on general-purpose methods—i.e., approaches based on loss functions that can be applied to standard network architectures without requiring prior knowledge about the data (such as semantically meaningful transformations) and that do not rely on any form of external dataset. Our goal is to evaluate MV-PMODE as a general-purpose density estimator on a benchmark where anomaly detection has been the subject of extensive research, thereby providing a meaningful test of its practical competitiveness. Under these constraints, the main applicable deep method families are one-class classification, autoencoders, and density estimation or generative models.

Prior to \cite{ruff18}, autoencoders and energy-based density estimators were the standard approaches to deep anomaly detection \citep{ruff21}. Energy-based models such as restricted Boltzmann machines are not particularly scalable, and to our knowledge, have not been applied to the CIFAR-10 AD benchmark. The paper \cite{ruff18} includes results for a deconvolutional autoencoder, which performs poorly on the classes where MV-PMODE achieves its strongest results. The only method on the CIFAR-10 AD leaderboard on \url{paperswithcode.com} that explicitly mentions an autoencoder is DASVDD \citep{dasvdd}, which we will discuss later.

Around 2017, new generative models such as variational autoencoders (VAEs), generative adversarial networks (GANs), and normalizing flows gained popularity. Models that directly estimate a density $\hp$ were found to perform poorly on anomaly detection tasks \citep{nalisnick18}—for example, reporting MNIST digits as more "normal" than in-class FMNIST images. Even normalizing flow-based methods that \emph{learn the pdf over features extracted from a pretrained network} do not perform especially well on CIFAR-10 AD; for instance, \cite{fastflow} achieves a mean AUC of 66.7 despite relying on a strong pretrained backbone.

Slightly better performance has been observed with GAN-based methods. AnoGAN and \emph{fast AnoGAN} \citep{anogan,fanogan} were among the earliest GAN-based approaches to anomaly detection. These methods train a GAN on nominal data—mapping a latent variable $z \sim \mathcal{N}(0, I)$ to the image space via $f_\theta(z)$—and then, at test time, minimize $\min_z \left\|\Xtest - f_\theta(z)\right\|_2^2$ to produce an anomaly score. However, these papers did not report results on CIFAR-10 AD. A closely related method, AD-GAN \citep{deecke19}, improves over f/AnoGAN and includes author-reported results on CIFAR-10, which we reproduced in the main text.

Many modern anomaly detection methods incorporate deep learning components—such as GANs or contrastive objectives—into increasingly elaborate pipelines. One such example is P-KDGAN \citep{pkdgan}, the best-performing GAN-based method we are aware of on CIFAR-10 AD. It employs eight networks in a knowledge distillation setup, combining a student–teacher architecture with multi-stage training and a weighted three-term loss. While it achieves a strong 73.8 mean AUC, its complexity and reliance on benchmark-specific design make it unsuitable as a general-purpose density estimator. Indeed, it is evaluated only on CIFAR-10 AD and the simpler MNIST AD benchmark. Nevertheless, we include it in Table~\ref{tab:additional-results}. Notably, even with its task-specific tuning, it does not outperform MV-PMODE on all classes.

Finally, we mention one-class classification methods for deep anomaly detection. Given nominal data $X_1,\ldots, X_n$, these approaches train a network to minimize the objective $\sum_{i=1}^n \|f_\theta(X_i) - c\|_2^2$ for a fixed center $c$, typically chosen at random. The earliest and most well-known example is Deep Support Vector Data Description (DSVDD), which we included in the main text. DSVDD uses a two-stage training procedure: first pretraining with an autoencoder loss to learn useful features, followed by training with the one-class loss above. This method was later extended to \emph{Deep Autoencoding Support Vector Data Descriptor} (DASVDD) \citep{dasvdd}, which jointly trains the autoencoder and one-class objectives in a unified way, yielding a modest improvement over DSVDD with a mean AUC of 66.5. We include DASVDD in Table~\ref{tab:additional-results}; notably, on the CIFAR-10 AD classes where MV-PMODE performs best, DASVDD does not outperform it. Other extensions of deep one-class classification rely on external datasets or pretrained backbones, making them unsuitable for comparison in our setting.

\begin{table}[ht]
    \centering \footnotesize
    \caption{Additional Results on CIFAR-10 One vs. Rest Benchmark}
    \label{tab:additional-results}
    \begin{tabular}{l*{11}{c}}
        Method      & Air.  & Auto. & Bird  & Cat    & Deer  & Dog   & Frog  & Horse & Ship & Truck & Mean\\ \hline
        DCAE\footnote{Values from \cite{ruff18}.}        & 59.1  & 57.4  & 48.9  & 58.4   & 54.0  & 62.2   & 51.2  & 58.6 & 76.8 & 67.3 &59.4\\
        P-KDGAN     & 82.5  & 74.4  & 70.3  & 60.5   & 76.5  & 65.2   & 79.7  & 72.3 & 82.7 & 73.5 &73.8\\
        DASVDD      & 68.6  & 64.3  & 55.8  & 58.6   & 64.0  & 62.6   & 71.0  & 64.6 & 81.1 & 73.7 & 66.5
    \end{tabular} 
\end{table}
Ultimately, we chose CIFAR-10 as a challenging, high-dimensional dataset where each class is highly multimodal and complex—providing a rigorous setting for performance evaluation. In the main text, we focused on two well-known and widely cited methods that use straightforward training procedures and do not rely on external datasets, pretrained backbones, or highly engineered architectures. We view these as fair baselines for MV-PMODE, which is a general-purpose density estimator not designed specifically for image data or anomaly detection.

\section{Anomaly Detection Experiment Details} \label{appx:ad-experiment}
\subsection*{Implementation Details}
For MV-PMODE, we used $k = 20$ mixture components, with $1200$ samples allocated to the estimation set and $3800$ to the validation set. Initial partition labels were generated using $k$-means clustering, followed by a hill-climbing optimization procedure.

Optimization proceeded by randomly selecting a subset of the labels (at rates described below), reassigning each to an independently sampled random class label, and accepting the change if it reduced the validation loss. If no improvement was found after 50 attempts, the perturbation size was reduced. The perturbation rates were 5\%, 2\%, 1\%, and 0.1\%. The procedure was parallelized to evaluate 10 random perturbations concurrently.

Each full optimization (i.e., one run for one CIFAR-10 class) was capped at 1800 seconds (30 minutes) to limit runtime. In nearly all cases, optimization continued until this time limit was reached. The complete experiment—five runs per class across ten classes—took just under two days to complete.

\subsection*{Hardware}
All experiments were run on a Mac Mini equipped with an Apple M2 Pro chip (10-core CPU, 32\,GB unified memory). The machine also has a 16-core GPU, which was not used for this experiment.

\subsection*{Full Results}
The table below reports the full results of our experiment, including all runs for each class.
\begin{table}[ht]
    \centering \footnotesize
    \caption{Additional Results on CIFAR-10 One vs. Rest Benchmark. Results in AUROC.}
    \label{tab:full_results}
    \begin{tabular}{l*{11}{c}}
        Class & Airplane & Auto. & Bird & Cat& Deer & Dog & Frog & Horse & Ship & Truck \\
        \hline
        Run 1 & 73.7 & 47.5 & 69.3 & 52.3 & 76.6 & 50.7 & 75.6 & 53.6 & 74.4 & 55.9 \\
        Run 2 & 74.3 & 48.2 & 68.5 & 49.7 & 76.9 & 50.6 & 75.1 & 55.8 & 74.6 & 54.5 \\
        Run 3 & 73.9 & 49.8 & 68.6 & 51.5 & 76.7 & 50.2 & 75.0 & 54.4 & 76.3 & 54.7 \\
        Run 4 & 73.9 & 48.7 & 68.7 & 51.5 & 76.5 & 51.3 & 75.5 & 53.9 & 76.2 & 51.5 \\
        Run 5 & 73.1 & 50.2 & 68.9 & 51.7 & 76.9 & 49.5 & 75.1 & 55.1 & 73.4 & 53.6 \\
        Mean & 73.8 & 48.9 & 68.8 & 51.3 & 76.7 & 50.5 & 75.3 & 54.6 & 75.0 & 54.0 \\
        Median & 73.9 & 48.7 & 68.7 & 51.5 & 76.7 & 50.6 & 75.1 & 54.4 & 74.6 & 54.5 \\
        Std. Dev.& 0.40 & 1.01 & 0.29 & 0.87 & 0.15 & 0.59 & 0.22 & 0.78 & 1.13 & 1.45 \\
        \hline
        DSVDD Std. Dev & 4.1 & 2.1 & 0.8 & 1.4 & 1.1 & 2.5 &2.6 &0.9 &1.2 &1.2
    \end{tabular} 
\end{table}
We briefly note that MV-PMODE exhibits lower standard deviation than DSVDD across all classes except \textit{truck}, where its variance is slightly higher. In some cases, the reduction in variance is an order of magnitude. Standard deviations were not reported in the original ADGAN paper.

\subsection*{Discussion}
We believe there is substantial room for improving MV-PMODE's performance on CIFAR-10 AD. The method’s hyperparameters—such as the number of components $k$ and the training/validation split—were chosen based on a few simple preliminary tests, without extensive tuning. Similarly, the bandwidth parameters $\sigma_i$ were selected using a basic heuristic (Silverman's rule) without further refinement.

Additionally, there exist robust variants of kernel density estimation, such as those described in \cite{rkde}, which have been shown to improve empirical performance. While applying such methods to high-dimensional KDEs or even the naive Bayes KDE used here may be computationally intensive, they could be applied independently to each marginal within each mixture component. 

Altogether, these considerations suggest that MV-PMODE could be significantly improved using robust, off-the-shelf enhancements—particularly with access to greater computational resources.
\subsection*{Software Credits}
These experiments—and the PMODE software framework—were implemented using Python 3.12.10. The following Python packages were used:
    \begin{itemize}
        \item Numpy 2.0.1 \citep{numpy}
        \item Scipy 1.14.0 \citep{scipy}
        \item Scikit-learn 1.6.0 \citep{sklearn}
        \item Numba 0.61.2 \citep{numba}
    \end{itemize}
\section{Supplementary Experiments} \label{appx:supp-experiments}
We present here some supplementary experiments using PMODE on Gaussian mixture models to give some idea how the methodology compares in simple mixture modeling. All experimental code and results are contained in the included code supplement.

We compare three density estimation pipelines:

\begin{enumerate}
\item \textbf{sklearn GMM:} The standard EM implementation from \texttt{scikit-learn} \citep{sklearn}, using \texttt{GaussianMixture(max\_iter=1000, n\_init=1)}. We set \texttt{n\_init=1} to match PMODE, which is also optimized from a single initialization. The \texttt{max\_iter} parameter was increased from its default value of 100 to 1000 to allow for a more thorough search, consistent with the fact that PMODE always runs to a local minimum. We refer to this baseline as \emph{GMM}.
    \item \textbf{$L^2$-PMODE:} A PMODE variant using $k$ univariate Gaussian components, optimized to minimize the empirical $L^2$ distance.
    \item \textbf{KL-PMODE:} A similar PMODE variant that instead minimizes empirical negative log-likelihood (i.e., KL divergence).
\end{enumerate}
\subsection{Diabetes Gaussian Mixture Experiment: PMODE vs. EM}

\label{appx:diabetes-gmm}

For each $k \in {2,\dots,8}$, we performed 30 independent runs. Each run was carried out as follows:

\begin{itemize}
    \item The Diabetes dataset was shuffled, and 350 samples were selected for training. The remaining samples were used for evaluation. Unlike the setting used in the CIFAR-10 experiment, the full training set was used for both partition optimization and validation, as the model has relatively few parameters and is unlikely to overfit.

    \item PMODE was initialized using $k$-means clustering applied to the training set. This mirrors the initialization used in \texttt{scikit-learn}'s GMM implementation.

    \item PMODE optimization used a greedy hill-climbing procedure: at each step, all single-label perturbations were considered (i.e., reassigning one point to a different component). Once a perturbation was found that reduced the loss, it was accepted and the process continued. Optimization terminated when no further single-point reassignment improved the objective—i.e., a local minimum was reached. Perturbations were evaluated in parallel, allowing multiple candidate updates to be tested simultaneously.

    \item Final performance was measured via average test log-likelihood:
        \[
            \frac{1}{|X_{\mathrm{test}}|} \sum_{x \in X_{\mathrm{test}}} \log \hat{p}(x),
        \]
        where $\hat{p}$ is the learned mixture density.
\end{itemize}
\subsubsection*{Results and Discussion}
\begin{figure}[ht]
    \centering
    \includegraphics[width=\linewidth]{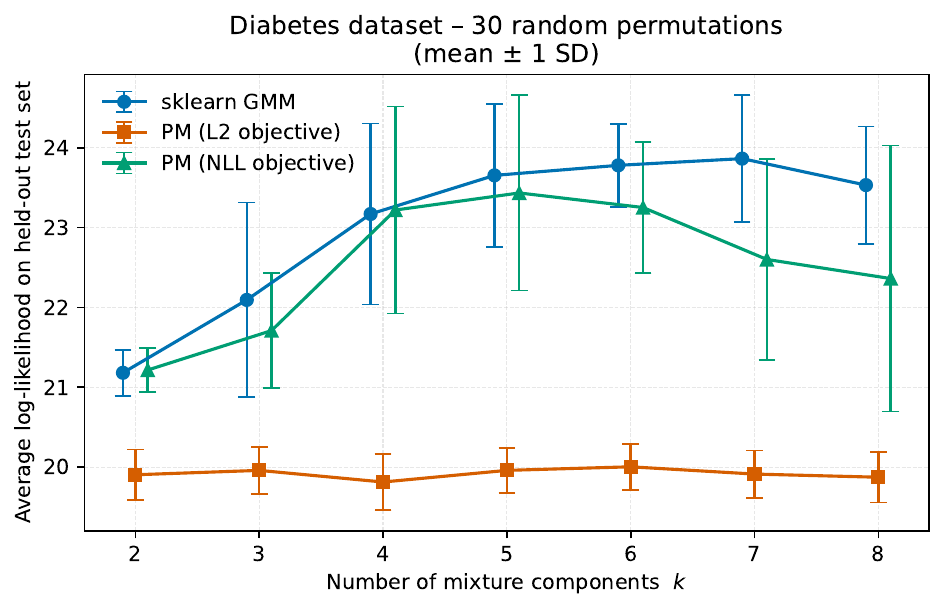}
    \caption{\textbf{Diabetes test log-likelihood.}
             Markers: means over 30 shuffles; error bars: \emph{standard
             deviation}.}
    \label{fig:diabetes-gmm}
\end{figure}

The experiment was run on the same machine described above and took approximately 3 hours to complete all 630 runs. Runs with larger $k$ values took noticeably longer, though not dramatically so. Figure~\ref{fig:diabetes-gmm} shows the mean and \emph{sample standard deviation} across the $30$ runs. Table \ref{tab:diab-wilcoxon} contains results for running the two-sided Wilcoxon signed-rank test on the runs of GMM vs KL-PMODE. This is a paired test to determine if the mean between paired samples is $0$, here each pair is one run with each estimator.

Figure~\ref{fig:diabetes-gmm} shows the mean and \emph{sample standard deviation} across the 30 runs. Table~\ref{tab:diab-wilcoxon} reports the results of a two-sided Wilcoxon signed-rank test comparing GMM and KL-PMODE. This is a paired test that checks whether the distribution of differences between the methods is centered at zero; in our case, each pair consists of a single run from each estimator.

Unsurprisingly, $L^2$-PMODE performs poorly in KL divergence across all values of $k$, reflecting a mismatch between training and evaluation objectives. KL-PMODE performs better than GMM when $k = 2$, with no significant difference observed for $k = 3, 4, 5$, after which GMM begins to outperform. Notably, KL-PMODE's performance peaks at $k = 5$ and declines thereafter, whereas GMM continues improving until $k = 7$ before tapering off.

The reason for KL-PMODE's earlier decline is unclear. One possibility is that, due to the discrete nature of the optimization problem, it may get stuck in poorer local minima as $k$ increases. Unlike EM, which benefits from a smooth objective and analytical updates, KL-PMODE may become more sensitive to local fluctuations in partitioning as model complexity grows. Additionally the model may simply lack the flexibility  when dividing 350 samples into 7 partitions. Perhaps the method would perform better if the "partitions" were allowed to be overlapping subsets—so that different mixture components could be estimated using shared data points, thus granting more flexibility.

\begin{table}[ht]
    \caption{
        Two-sided Wilcoxon signed-rank test evaluating the statistical significance of per-run test log-likelihood differences between KL-PMODE and GMM for the \textbf{Diabetes} dataset. The first row reports the median difference across 30 runs (larger values generally favor PMODE). The second row reports the corresponding p-values. We omit $L^2$-PMODE from this comparison, as it consistently performed substantially worse in this experiment.
}
    \begin{tabular}{lccccccc}
        Number of components& 2         & 3             & 4         & 5         & 6             &7              & 8\\
        \hline
        median of differences & 0.028 & 0.058& 0.117 & -0.107 & -0.551 & -1.166 & -0.915\\
         p-val     & 4e-3 & 0.789   & 0.42 & 0.33 & 1.1e-3   & 3.8e-6   & 4.4e-5
     \end{tabular} \label{tab:diab-wilcoxon}
\end{table}
\newpage
\subsection{Iris Gaussian Mixture Experiment: PMODE vs. EM}

This experiment followed the same procedure as the previous one, but was conducted on the \textbf{Iris} dataset. We used 120 samples for training and cross-validation, with the remaining 30 used for evaluation. The experiment was run for $k \in {2, 3, 4, 5}$, with 30 independent runs for each value of $k$. All other experimental details were identical to those described in the Diabetes experiment.

\subsubsection*{Results and Discussion}

\begin{figure}[ht]
    \centering
    \includegraphics[width=\linewidth]{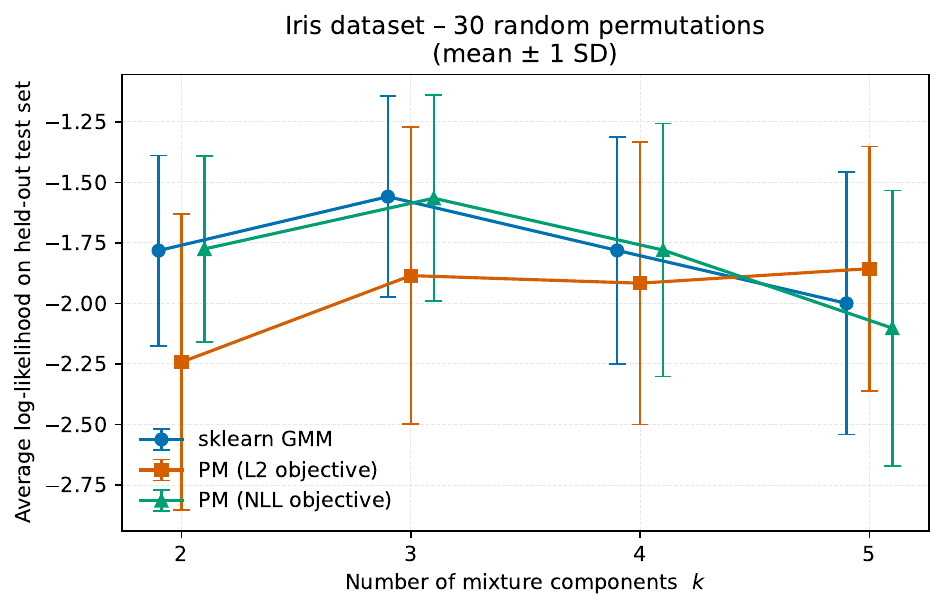}
    \caption{\textbf{Iris test log-likelihood.}
             Markers: means over 30 shuffles; error bars: \emph{standard
             deviation}.}
    \label{fig:iris-gmm}
\end{figure}

The experiment was run on the same computer as above and took approximately 5 minutes to complete all 120 runs. Figure~\ref{fig:iris-gmm} shows the mean and \emph{sample standard deviation} across the 30 runs. Table~\ref{tab:iris-wilcoxon} reports the results of the two-sided Wilcoxon signed-rank test comparing GMM to both KL-PMODE and $L^2$-PMODE. This paired test checks whether the average difference between matched runs is significantly different from zero.

We find that KL-PMODE performs nearly identically to GMM across most settings, with very small differences in test log-likelihood and generally low statistical significance. However, KL-PMODE does outperform GMM at $k = 2$ with high statistical significance, which is consistent with results from the Diabetes experiment.

The performance of $L^2$-PMODE is more variable. Interestingly, its relative performance appears to improve as $k$ increases, eventually outperforming both KL-PMODE and GMM. The reason for this trend is unclear. One possible explanation is that $L^2$-PMODE, by minimizing a squared-error criterion, may be more robust in regions where log-likelihood-based losses behave erratically near zero.
\begin{table}[ht]
    \centering
    \caption{
Two-sided Wilcoxon signed-rank test evaluating the per-run test log-likelihood differences between PMODE and GMM on the Iris dataset. For each $k$, the first and third rows show the median difference across 30 runs for KL-PMODE and $L^2$-PMODE, respectively (larger values generally favor PMODE). The second and fourth rows show the corresponding p-values.}
    \begin{tabular}{lcccc}
        Number of components& 2         & 3             & 4         & 5 \\
        \hline
        median of differences KL-PMODE& 0.0057 & 0.0092& -0.020 & -0.036 \\
         p-val     & 7e-5 & 0.40   & 0.87 & 0.39\\
         \hline
        median of differences $L^2$-PMODE& -0.454 & -0.250& 0.002 & 0.215 \\
         p-val     & 3e-4 & 2e-3   & 0.36 & 0.073
     \end{tabular} \label{tab:iris-wilcoxon}
\end{table}

\end{document}